# AI, Entrepreneurs, and Privacy: Deep Learning Outperforms Humans in Detecting Entrepreneurs from Image Data

**Martin Obschonka[a,b,*], Christian Fisch[c], Tharindu Fernando[d], Clinton Fookes[d]**

[a] *Amsterdam Business School, University of Amsterdam, Amsterdam, 1018 TV, The Netherlands*

[b] *Australian Centre for Entrepreneurship Research, Queensland University of Technology, Brisbane, 4000, Australia*

[c] *Interdisciplinary Centre for Security, Reliability and Trust (SnT), University of Luxembourg, Luxembourg, 1855, Luxembourg*

[d] *Faculty of Engineering, Queensland University of Technology, Brisbane, 4000, Australia*

* Corresponding author.

*E-mails:* m.obschonka@uva.nl (M. Obschonka), christian.fisch@uni.lu (C. Fisch), t.warnakulasuriya@qut.edu.au (T. Fernando), c.fookes@qut.edu.au (C. Fookes)

**Abstract:** Occupational outcomes like entrepreneurship are generally considered personal information that individuals should have the autonomy to disclose. With the advancing capability of artificial intelligence (AI) to infer private details from widely available human-centric data (e.g., social media), it is crucial to investigate whether AI can accurately extract private occupational information from such data. In this study, we demonstrate that deep neural networks can classify individuals as entrepreneurs with high accuracy based on facial images sourced from Crunchbase, a premier source for entrepreneurship data. Utilizing a dataset comprising facial images of 40,728 individuals, including both entrepreneurs and non-entrepreneurs, we train a Convolutional Neural Network (CNN) using a contrastive learning approach based on pairs of facial images (one entrepreneur and one non-entrepreneur per pair). While human experts (n=650) and trained participants (n=133) were unable to classify entrepreneurs with accuracy above chance levels (>50%), our AI model achieved a classification accuracy of 79.51%. Several robustness tests indicate that this high level of accuracy is maintained under various conditions. These results indicate privacy risks for entrepreneurs.

## 1. Introduction

The ability of artificial intelligence (AI) systems, such as deep learning models, to extract complex and valid information from digital traces has emerged as a pivotal research topic across various scientific disciplines. As AI evolves, their applications in analyzing diverse types of digital traces—such as facial images, social media text, medical records, and other forms of digital footprints—have expanded exponentially. Recent advancements in AI, for instance, leverage facial recognition technology to infer a range of basic personal attributes (e.g., gender, age, smile) from facial images [1]. Furthermore, an increasing number of studies demonstrate that AI can also identify more intimate personal attributes from facial images with accuracy significantly higher than humans. For example, research reports above-human accuracy for attributes like personality [2], sexual orientation [3], and political orientation [4, 5]. These developments raise significant ethical concerns related to privacy, civil liberties, and AI's overall predictive power [4, 5, 6, 7], highlighting the need to better understand the extent to which AI methods can detect private information in widely shared *human-centric data*, as well as the associated ethical risks.

To contribute to this debate, we extend research on AI's ability to capture personal attributes to the hitherto unexplored occupational domain. Specifically, we explore whether deep learning can distinguish between entrepreneurs and non-entrepreneurs from their facial images. We use a specific dataset drawn from Crunchbase (www.crunchbase.com), a premier data source for entrepreneurship data that is commonly used by researchers and practitioners. We also evaluate the accuracy and robustness of this classification and compare it to human classification performance.

We focus on distinguishing between entrepreneurs and non-entrepreneurs for several reasons. First, this dichotomy represents a well-established higher-level categorization of occupations [8, 9] that significantly influences an individual's occupational choices. Second, although many individuals voluntarily disclose such occupational information (e.g., on social media platforms like LinkedIn or company websites), the information remains fundamentally private. So far, it is unclear to what extent this private occupational information can be inferred from widely shared human-centric data, particularly when individuals are unaware that they might be revealing private details, using ubiquitous AI methods. Third, entrepreneurs and their ventures play a critical role in the economy, particularly in innovation and job creation [10, 11], making them important subjects for research, education, and policymaking. Last, there is considerable interdisciplinary research interest in private personal attributes as correlates of entrepreneurship [12, 13, 14, 15, 16].

Note that we deliberately refrain from interpreting the data regarding systematic facial differences—our primary concern lies with the significant ethical issues highlighted by the broader scientific and scholarly ethics community [6, 17, 18]. Thus, we solely focus on the deep learning's accuracy when classifying entrepreneurs vs. non-entrepreneurs. We also acknowledge that our dataset is not fully representative of the general population and may inadvertently transfer existing social biases and discrimination into AI results, potentially reinforcing or even amplifying such biases. For example, social biases and systemic discrimination could influence who pursues and succeeds in entrepreneurial careers, thereby shaping the available data. Still, when interpreted most ethically and rigorously, our findings provide valuable insights into the power of AI and its potential social, ethical, and methodological implications. As such, and like related research [2, 3, 4, 5], we deem it important to carry out and report on such research to inform the public about the power of AI, as well as its potential social, ethical, and methodological ramifications, thereby contributing to the ongoing discussions.

## 2. Method

All studies reported are approved by the Research Ethics Committee of the Queensland University of Technology (approval no. 2000000651) and the Economics and Business Ethics Committee of the University of Amsterdam (approval no. EB-6781; additional analysis with trained participants) and we strictly follow all relevant guidelines and regulations.

### 2.1 AI Model: Data, Design, and Training

#### 2.1.1 Data Collection

Our sample of facial images comes from Crunchbase (www.crunchbase.com), one of the premier databases in contemporary entrepreneurship research [14, 19, 20]. Importantly, Crunchbase provides information on entrepreneurs and non-entrepreneurs (e.g., managers and other employees of entrepreneurial ventures, investors). Each individual has a publicly accessible profile that contains demographic information, employment and entrepreneurship history, and profile pictures (mostly facial images). Our sample was recorded in March 2019 and includes individuals with a CB rank (Crunchbase's internal identifier indicating prominent individuals) between 1 and 100,000. W in the US with non-missing gender information. This yielded an initial sample of 42,043 individuals. Individuals who report having founded at least one ventures are labeled as entrepreneurs (n=25,071, 59.6%). The sample is reduced to 40,728 after removing individuals with invalid facial images (e.g., placeholders, comic images). Females constitute 19% of our sample.

Facial images only recently gained attention in entrepreneurship research, as a source of information about emotions [22, 23], attractiveness, competence, and intelligence [24]. Additionally, a recent study analyzed facial appearance and geometry (i.e., facial width-to-height ratio, cheekbone prominence, facial symmetry) to predict whether an individual emerges as an entrepreneur and entrepreneurial success [25].

### 2.1.2 Data Pre-processing

As facial images can contain information apart from the individual's face (e.g., background, other body parts) we first employ a face detector [28] to localize the face region in each image. The face detector outputs face coordinates, which we use to crop the image so that only the face region remains. We then resize the cropped facial images to a size of 224×224 pixels to match the input requirements of the pre-trained feature extraction model we employ later.

### 2.1.3 Entrepreneurship Classifier Design and Training

To develop our AI model for distinguish entrepreneurs from non-entrepreneurs based on facial characteristics, we adopt a two-stage approach.

First, we employ VGGFace2 [26], a popular deep learning model pre-trained on a large-scale dataset of facial images, to extract robust facial features. In our pipeline, each pre-processed facial image is passed through a shared VGGFace2 backbone, ensuring that all images are mapped into a common feature space. Deep learning models like VGGFace2 extract informative features from raw data by learning hierarchical representations: lower layers capture basic elements such as edges and textures, while higher layers combine these to form more abstract features such as object shapes. Each layer in the model contains thousands of trainable weights that are optimized during training to capture relevant patterns. Because tuning thousands of trainable weights from scratch requires vast amounts of data, transfer learning helps overcome this bottleneck by leveraging models originally trained on large datasets—an approach commonly used in face analysis to detect distinct facial characteristics.

Next, we fine-tune a Convolutional Neural Network (CNN) classifier on these extracted features. The objective of this CNN is to refine the pre-extracted features and highlight those most relevant to our classification task. More specifically, we employ a contrastive learning approach that compares pairs of facial images. The contrastive framework enables the model to learn by emphasizing both the similarities and differences between the two faces, a strategy particularly effective in tasks like face verification. Figure 1 provides an overview of our contrastive classification model.

*- Please insert Figure 1 about here -*

Our approach differs from other approaches used for facial recognition tasks related to latent personal attributes [3, 5]. For example, [5] extract VGGFace2 feature vectors and directly classify them using a logistic regression model. Similarly, [3] extract VGGFace features, apply a dimensionality reduction technique, and then classify the reduced features using logistic regression. The primary aim in both studies is to demonstrate that large-scale pre-trained facial recognition models (e.g., VGGFace and VGGFace2), inherently capture latent personal like political and sexual orientation during their training. In contrast, our goal is to design a deep learning model that accurately predicts entrepreneurial characteristics from facial images. To achieve this, we fine-tune the VGGFace2 features using a sophisticated contrastive learning-based classifier. This approach not only refines the feature representations for our specific task but also leverages the strengths of contrastive learning to better discriminate between entrepreneurs and non-entrepreneurs. This is a common approach in previous research [47, 48].

During training, we provide pairs of facial images to our model. Each pair consists of one entrepreneur and one non-entrepreneur. We ensure that each person in our sample is only included once in our pairs of images. We label the images within each pair as the left image and the right image, ensuring that both images come from individuals of the same gender. We randomize the position of the entrepreneur's facial image, so it can either appear as the left image or as the right image. Each image pair is treated as a single instance with one corresponding prediction. The model outputs a binary decision for each pair: 0 if it predicts that the entrepreneur is the left image, and 1 if it predicts that the entrepreneur is the right image. This approach ensures there is no double-counting of predictions. Through this learning process, the model learns which features are important for distinguishing between the two categories and can accurately determine which image belongs to an entrepreneur.

Following standard ML protocols, we randomly select 75% of these pairs for model training and the remaining 25% for model testing. From the training data, we hold out 10% as a validation set to evaluate the model on unseen data. Good validation performance suggests that the model generalizes well without overfitting. To account for variability in model initialization, we randomly initialize the model's weights 10 times.

Figure A1 (Appendix A) displays the training and validation accuracy curves. The similar performance between these curves indicates that the model is not overfitting.

### 2.2 Experiment with Human Entrepreneurship Experts

We compare the accuracy of our AI model with human entrepreneurship experts. We design a survey-based online experiment that mirrors the classification task of our AI model. The main part of the survey shows a consecutive set of pairs of facial images to participants (one entrepreneur and one non-entrepreneur per pair). Participants are asked to indicate which one is the entrepreneur. The Facial images in our online experiment comprise approximately 2,150 male and 500 female individuals randomly selected from our test set (i.e., these images are not used to train the AI model).

Each participant is confronted with 10 pairs of facial images. If a participant correctly identifies the entrepreneur in 10 out of 10 image pairs, their accuracy is 100%, while random guessing would result in an accuracy of 50%. Because our sample contains famous individuals, we also ask respondents to tick a box if they recognize any one or both of the individuals per pair. When calculating accuracy, we remove pairs where respondents indicated familiarity with the depicted individuals.

We also collect participants' age, country/region of origin, gender, the highest degree of education, the primary field of education, occupation, and expertise((a) full-time entrepreneur, (b) part-time entrepreneur, (c) professional investor (i.e., venture capitalist or business angel), (d) other employees, (e) entrepreneurship researcher, (f) entrepreneurship educator, (g) student, (h) prefer not to say, (i) none of the above. Because we are interested in the performance of entrepreneurship experts, we only keep participants that self-assign to (a), (b), (c), (e), or (f).

We solicit participants in two ways: First, we used Prolific (www.prolific.com) to recruit a sample of entrepreneurs. Second, we recruit participants by distributing the survey in the authors' professional networks and via social media (addressing entrepreneurship circles). In total, we collect responses from 650 (self-assigned) entrepreneurship experts who classify 6,500 face-pairs. Because we remove all decisions in which the respondents indicated familiarity with the individuals depicted, our analyses consider 6,431 out of 6,500 classification tasks.

Table 1 displays a breakdown of our human entrepreneurship experts according to expert category, origin, age, gender, highest education degree, and primary field of education.

*- Please insert Table 1 about here -*

## 3. Main Results

### 3.1 AI Model's Accuracy

The average accuracy of the AI model is obtained from randomly initializing the internal weights of the AI model and training the model 10 times. The accuracy of the model, when it converges, is taken as the accuracy of that trial. The accuracies obtained in the 10 trials are [80.30, 78.76, 79.28, 77.89, 79.98, 79.53, 79.52, 80.39, 79.20, 80.24], which yield an average accuracy of 79.51 (SD=0.78). This means that when presented with a pair of images, our AI model can identify the entrepreneur 79.51% of the time well above random guessing (i.e., 50%). This indicates that our AI model can identify systematic differences that distinguish the facial images of entrepreneurs from those of non-entrepreneurs.

## 3.2 Accuracy Achieved by Human Entrepreneurship Experts

The mean accuracy across all subgroups of human experts is 49.42% (SD=15.93; see Table 2). The accuracy is highest among entrepreneurship researchers (51.24%), and lowest among professional venture capitalists and business angels (43.87%). However, these differences are not very pronounced: the mean accuracy of the human judges (49.42%) is relatively homogeneous around or slightly below the mean value of 50% or a random guess, indicating that human experts cannot systematically distinguish entrepreneurs from non-entrepreneurs. The result of the t-test in Table 2 documents that our human entrepreneurship experts achieve significantly lower accuracies than the AI model ($p<0.01$), indicating that the AI model's performance is "beyond-human".

*- Please insert Table 2 about here -*

## 3.3 Robustness Checks

To shed some light on the functioning of the AI model, our findings' robustness, and their validity, we perform a range of further analyses and robustness checks. We briefly summarize these analyses below and report more technical details in Appendix A.

### 3.3.1 Exploring the AI model's decisions

We observe considerable variability in the visual information that our AI model uses to classify entrepreneurs and non-entrepreneurs. To illustrate, Figure 2 presents heatmaps for two facial image pairs of entrepreneurs and non-entrepreneurs (pair 1: images a and b, pair 2: images c and d), highlighting the areas in the input images critical for classification. Each heatmap shows the top 50 sub-regions contributing to the decision, with the sub-region boundaries indicated in yellow.

Considering this variability in the selected sub-regions, we systematically analyze using standard central facial landmarks (i.e., nose, eyes, mouth). In this analysis, we provide only a single facial region or a combination of regions as input to determine which of them are most significant for distinguishing entrepreneurs from non-entrepreneurs (for details, see Appendix A2). The results in Figure 3 show that the visual information associated with the nose region yields the highest accuracy. Additionally, accuracy improves when the information from multiple facial landmarks is considered jointly. However, even the best-performing landmark-based model does not outperform our main model, which uses the entire facial image as input, indicating that our main model can capture other important facial attributes (e.g., skin textures) and systematically discovers non-linear relationships among them.

*- Please insert Figures 2 and 3 about here -*

### 3.3.2 Model Bias: Gender and Race

Because AI models are prone to biases, we explicitly analyze the sensitivity of our trained model to gender and race.

Given that most of our training data is from male individuals (81%), we first assess whether our model's accuracy differs between male and female images. On the test set, our model achieves an accuracy of 78.5% for male images and 83.1% for female images. These results suggest that the model performs comparably when identifying male and female entrepreneurs. To further investigate potential gender bias, we generate an embedding space visualization for a set of randomly selected samples (Appendix A3). This analysis indicates that the model's learned embedding space differentiation between entrepreneur (ENT) and non-entrepreneur (Non-ENT) labels and without bias toward a specific gender.

Racial bias is another area of concern, given that most individuals in our sample are white. This imbalance could potentially cause our model to perform differently on facial images of non-white individuals. Because racial background information is not available in Crunchbase, two authors inspected all the ~2,600 facial images used in our online experiment, to identify non-white individuals. We then separately test the trained model using these subsets of non-white images identified by each researcher. The resulting accuracies for the non-white images identified by researcher 1 are 78.24% for male individuals (n=489) and 80.75% for female individuals (n=140). Similarly, the accuracies for the non-white images for researcher 2 are 77.90% for male individuals (n=596) and 81.84% for female individuals (n=176). These results are consistent with our main findings, indicating that the AI model does not appear to be heavily biased toward a particular racial background in our dataset.

### 3.3.3 Accuracy Achieved by Trained Human Participants

While our AI model undergoes a training process, our human entrepreneurship experts do not receive any training. In this robustness check, we therefore examine whether such training would have an effect.

We devise a brief training program by extracting a random sample of image pairs from our online classification experiment (48 male pairs, 12 female pairs, in line with the gender distribution in our full sample). We prepare a presentation (PowerPoint slides), including 12 entrepreneurs and 12 non-entrepreneurs per slide (=24 facial images per slide). These images are labeled as 'entrepreneur' or 'non-entrepreneur'. We use these training slides (see Appendix B) in a classroom in entrepreneurship and business bachelor's and master's courses at the University of Amsterdam and the University of Luxembourg. Using the large screen, we expose students to the training material for approximately 10 minutes, asking them to concentrate on the images to examine and memorize any group differences. We then ask them to take our online classification experiment. As described in Table 2, we collected responses from 133 individuals, making 1,273 classifications. The average accuracy is 48.12% (SD=17.99) indicating that training does not significantly improve human performance.

Figure 4 summarizes our core results on the accuracy of the AI model versus human experts (main result) and versus the trained experts (robustness check).

*- Please insert Figure 4 about here -*

## 4. Discussion

While the findings of any single study should be approached with caution, our research indicates that deep learning algorithms can distinguish between entrepreneurs and non-entrepreneurs from publicly available, human-centric datasets with above-chance accuracy (79.51%). In contrast, human raters perform at chance levels. This adds to our knowledge of the capabilities of AI in (a) extracting a whole range of private personal information from readily accessible human-centric data [2, 3, 4, 5] and (b) outperforming humans, including experts, in such tasks. As highlighted by [4], "one's face is particularly difficult to hide in both interpersonal interactions and digital records," implying that private information derived from facial images—including occupational details—can be highly sensitive and easily disseminated among society, businesses, organizations, and individuals. For example, entrepreneurial experience can have value as information because it has been linked to advantages in terms of access to entrepreneurial resources in (budding) entrepreneurs, including venture capital [36, 37], increased venture survival [38], and public policy schemes [39, 40].

Although these findings are important, it is crucial to emphasize that their implications and applications must be approached with utmost caution. The risks are high and not fully understood. These risks extend beyond social bias, such as stereotypes and discrimination, to include a range of societal risks. Misapplication of these models can lead to significant ethical issues and negative impacts on society [6, 17, 18]. Therefore, our results should not be interpreted as an endorsement for the widespread use of such AI methods, including facial recognition, for evaluating and classifying individuals. The sole aim of this study was to assess

AI's capabilities compared to human performance within the limitations of the study design and potential social bias, which is a crucial consideration. For example, certain groups could be exposed to discrimination and stereotyping which could affect their likelihood of becoming an entrepreneur and hence representation in the dataset we used and the respective classification results. Social bias in the real world is well documented in a myriad of studies (e.g., in investment decisions affecting entrepreneurs [13] or the 'what is beautiful is good' stereotype [41]; see also [25]). We also acknowledge the well-documented dangers of following any 'illusions' of understanding AI-driven research results [42]. Thus, we present our findings by focusing on the predictive capacity of AI in comparison to human performance while also discussing these results in the context of potential biases and their broader societal implications. We believe that our findings and this contextualized interpretation hinting at the potential of social bias add significant knowledge to the respective debates in society—given the unquestionable disruptive potential of such AI methods and data in the real world, along with their major ethical implications affecting large parts of society.

### 4.1 Limitations

Our study has various limitations. First, although we conducted several additional tests, we did not study what information the AI model used to distinguish entrepreneurs from non-entrepreneurs. For example, there is chance that facial expression, head posture, lighting, or other situational factors could have played a role (see also [5]). Maybe the AI model is able to detect a certain confident look not detectable to humans (research found that entrepreneurs often differ from non-entrepreneurs in higher overconfidence). Perhaps the AI model can detect a certain confident expression that is not discernible to humans. Research has shown that entrepreneurs often exhibit higher levels of overconfidence compared to non-entrepreneurs. However, as noted earlier, our primary goal was not to identify the specific distinguishing features, but rather to test whether publicly available image data from entrepreneurs and non-entrepreneurs disclose private occupational information via AI (in comparison to human performance).

Second, as with any empirical research project, the quality of the data used to train and test the AI model is critical. If the data is inaccurate or biased, the AI model will absorb these inconsistencies and generate flawed conclusions that may reproduce or amplify the biases present in the training data [43, 44]. We use Crunchbase to retrieve and identify our sample of entrepreneurs and non-entrepreneurs. Crunchbase is one of the most ubiquitous databases used in contemporary entrepreneurship research due to its recent, accurate, and comprehensive coverage [14, 19, 20, 21]. Despite its coverage, we acknowledge that the founding history recorded in Crunchbase might not be completely accurate for every individual. For example, some individuals that we classify as non-entrepreneurs might have participated in founding a new venture that is too insignificant to be recorded in Crunchbase. Others might deliberately omit founding information from their profile to masquerade past failures. This leads to a situation in which we might falsely classify some individuals as non-entrepreneurs even though they are entrepreneurs. Future research could try to circumvent this potential limitation by collecting training data from different sources (e.g., via surveys) or by performing thorough background checks on the individuals included in Crunchbase to verify the accuracy of their classification as entrepreneurs or non-entrepreneurs.

We also acknowledge that the data included in Crunchbase that we use in our analyses might not to be free of bias or representative of the entire population of entrepreneurs and non-entrepreneurs. Specifically, the individuals that we classify as non-entrepreneurs represent prominent business professionals (e.g., CEOs, managers, investors) so that our sample of non-entrepreneurs is not a cross-section of the general population. Instead, our assessment is closer to a comparison between entrepreneurs and managers, which is a popular comparison in entrepreneurship research [45, 46]. Moreover, potential biases in our data could stem from the fact that Crunchbase focuses mostly on tech ventures [14] in the US [19]. This suggests that our sample might be skewed towards entrepreneurs and non-entrepreneurs in the US tech sector. While we acknowledge that this sample might not be representative of the entire population of entrepreneurs, we want to emphasize that such ventures are a particularly important source of economic growth and innovation [14]. To summarize, while Crunchbase is a state-of-the-art database in contemporary entrepreneurship research and we are not aware of an alternative data source that would allow us to improve our model (i.e., from a technical, legal, and ethical standpoint), the caveats that we acknowledge need to be considered when interpreting the results.

Third, our model is trained with one facial image per person (via the image pairs). Using more images per person could change the effectiveness of the training. It might also make it possible to study the role of age. While our sample covers individuals across all age groups, we do not know how the algorithm behaves for facial images of the same individual over time. That is, there might be some bias in our predictions related to age.

Finally, in our analysis, we attempted to address a potential bias regarding gender and racial background in the AI model but that cannot completely rule out any biases. Examples include potential gender identity, disabilities, or ancestry. Because we are unable to reliably infer these characteristics from the data available in Crunchbase, we cannot consider them in our assessment.

### 4.2 Conclusion

Together with a growing body of related findings, studies like ours show the power of AI and underscore the need for robust ethical guidelines and regulatory frameworks to govern the use of AI and human-centric data. This includes ensuring that the extraction of personal information from publicly available sources does not lead to misuse, compromise individual privacy, or violate ethical standards—particularly given the potential social biases inherent in such data and the resulting AI outputs. This also highlights the necessity for extreme caution regarding ethical risks in study designs, results, and their interpretation and application. As technology advances, it becomes imperative to balance its potential benefits with the deep ethical challenges it presents, ensuring that AI deployment respects individual privacy and aligns with societal values and ethical standards.

**Author contributions statement**

Martin Obschonka, Christian Fisch, Tharindu Fernando, and Clinton Fookes designed the research; Clinton Fookes developed the deep ML research idea and provided supervision; Martin Obschonka, Christian Fisch, and Tharindu Fernando performed the research and analyzed the data; Martin Obschonka, Christian Fisch, and Tharindu Fernando wrote the manuscript. All authors discussed the results. All authors reviewed and approved the manuscript.

**Competing interests statement**

The authors declare no competing interests.

**Data availability statement**

The data and code used in this study contain sensitive information and cannot be publicly shared. However, the code and data (except for the facial images) may be shared by the corresponding author upon reasonable request.

## Tables

**Table 1.** Background information about the human entrepreneurship experts (n = 650) who participated in our online experiment.

| Item | Category | Counts | Percent |
|---|---|---|---|
| **Expert group** | Entrepreneur | 384 | 59.08 |
| | Entrepreneurship educator | 92 | 14.15 |
| | Entrepreneurship researcher | 143 | 22.00 |
| | Venture capitalist/business angel | 31 | 4.77 |
| **Region** | Australia/Asia-Pacific | 145 | 22.31 |
| | Europe | 120 | 18.46 |
| | Middle East/North Africa | 10 | 1.54 |
| | South America | 11 | 1.69 |
| | USA/Canada | 355 | 54.62 |
| | No response | 9 | 1.38 |
| **Age** | 24 or younger | 40 | 6.15 |
| | 25 to 29 | 71 | 10.92 |
| | 30 to 39 | 195 | 30.00 |
| | 40 to 44 | 161 | 24.77 |
| | 50 to 54 | 123 | 18.92 |
| | 60 or older | 59 | 9.08 |
| | No response | 1 | 0.15 |
| **Gender** | Female | 244 | 37.54 |
| | Male | 375 | 57.69 |
| | Other | 3 | 0.46 |
| | No response | 28 | 4.31 |
| **Highest degree** | Below high school degree | 5 | 0.77 |
| | High school degree or equivalent | 87 | 13.38 |
| | Bachelor degree | 174 | 26.77 |
| | MBA | 45 | 6.92 |
| | Other Master degree/postgraduate | 137 | 21.08 |
| | PhD/doctoral degree | 182 | 28.00 |
| | Prefer not to say/no response | 20 | 3.08 |
| **Field of education** | Business or economics | 274 | 42.15 |
| | Humanities | 31 | 4.77 |
| | Law | 18 | 2.77 |
| | STEM | 159 | 24.46 |
| | Social sciences | 76 | 11.69 |
| | Other | 73 | 11.23 |
| | No response | 4 | 0.62 |

**Table 2.** Main results: mean accuracy of selected subgroups within our human judges and comparison with the performance of our AI model.

| Model/sample | n (classifications) | Mean accuracy (SD) | t-test Human experts vs. AI model (p-value) |
|---|---|---|---|
| **AI model** | **10 (-)** | **79.51 (0.78)** | **-** |
| | | | |
| **Human experts** | **650 (6,431[a])** | **49.42 (15.93)** | **5.92 (0.00)** |
| Entrepreneur | 384 (3,791) | 50.27 (15.66) | 5.90 (0.00) |
| Entrepreneurship educator | 92 (911) | 47.74 (17.35) | 5.76 (0.00) |
| Entrepreneurship researcher | 143 (1,419) | 51.24 (15.42) | 5.78 (0.00) |
| Venture capitalist/business angel | 31 (310) | 43.87 (14.30) | 7.81 (0.00) |
| | | | |
| **Trained humans** | **133 (1,273)** | **48.12 (17.99)** | **5.50 (0.00)** |

*Notes:* The human experts (n=650) did not undergo any specific training. The "trained humans" (n=133) were exposed to a brief training before participating in our online experiment. The t-test confirms a significant difference between the performance of human judges and AI model ($p < 0.01$). [a] = each classification refers to a human expert being shown a pair of facial images and indicating who they think is an entrepreneur. While every respondent performed 10 classifications, the number of classifications that we use to calculate the accuracies is slightly lower than 6,500 (=650 participants making 10 classifications each) because we remove those classifications in which respondents indicated that they know one of the facial images (i.e., recognized the entrepreneur or non-entrepreneur).

## Figures

**Figure 1:** Architecture of the CNN-based classification model.

**Figure 2:** Example visualizations highlighting the most salient sub-regions for the model decision (in red) and sub-region boundaries (in yellow) from individual cases.

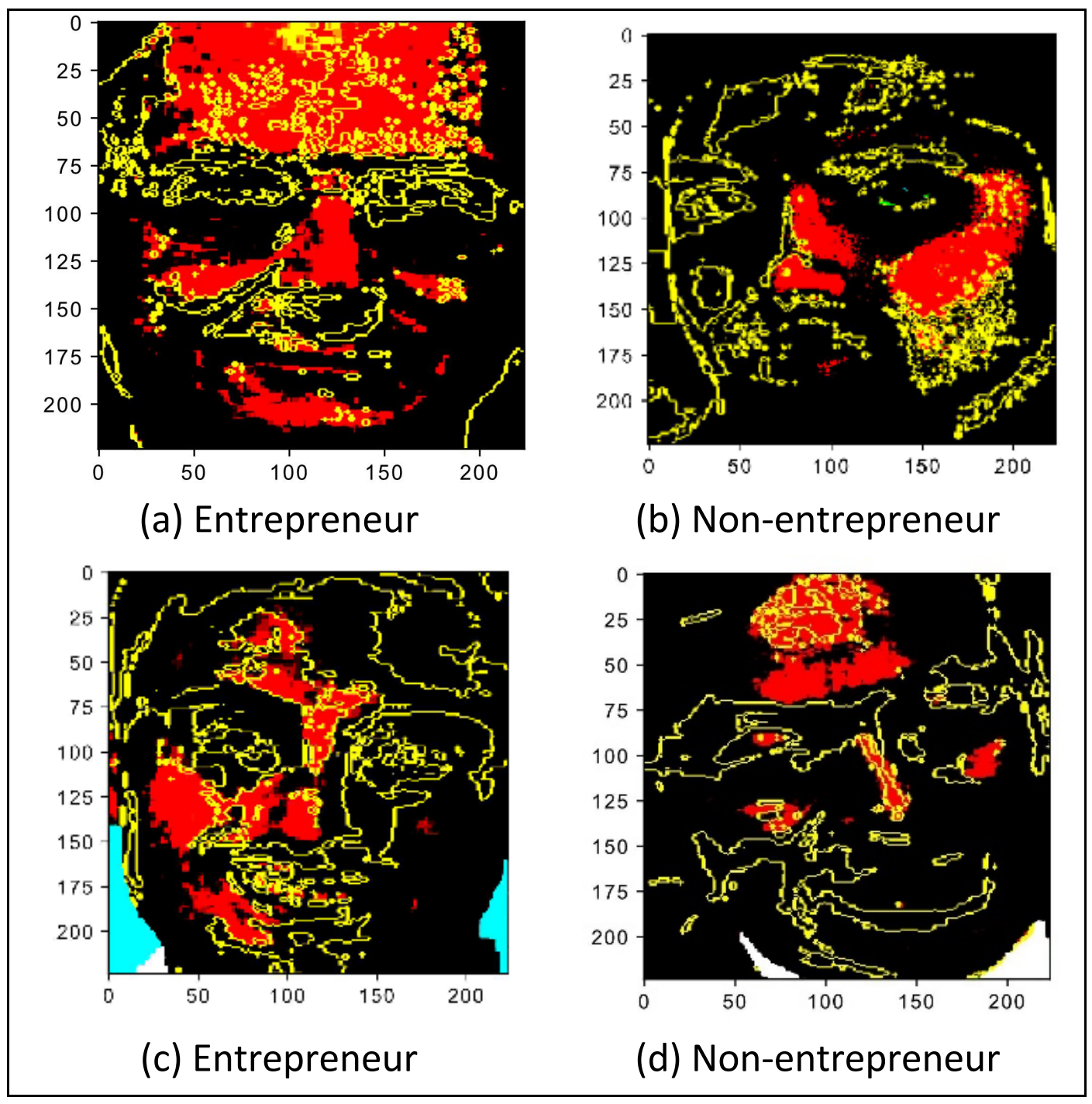

**Figure 3:** Results (accuracy) of the facial landmarks-based classifier.

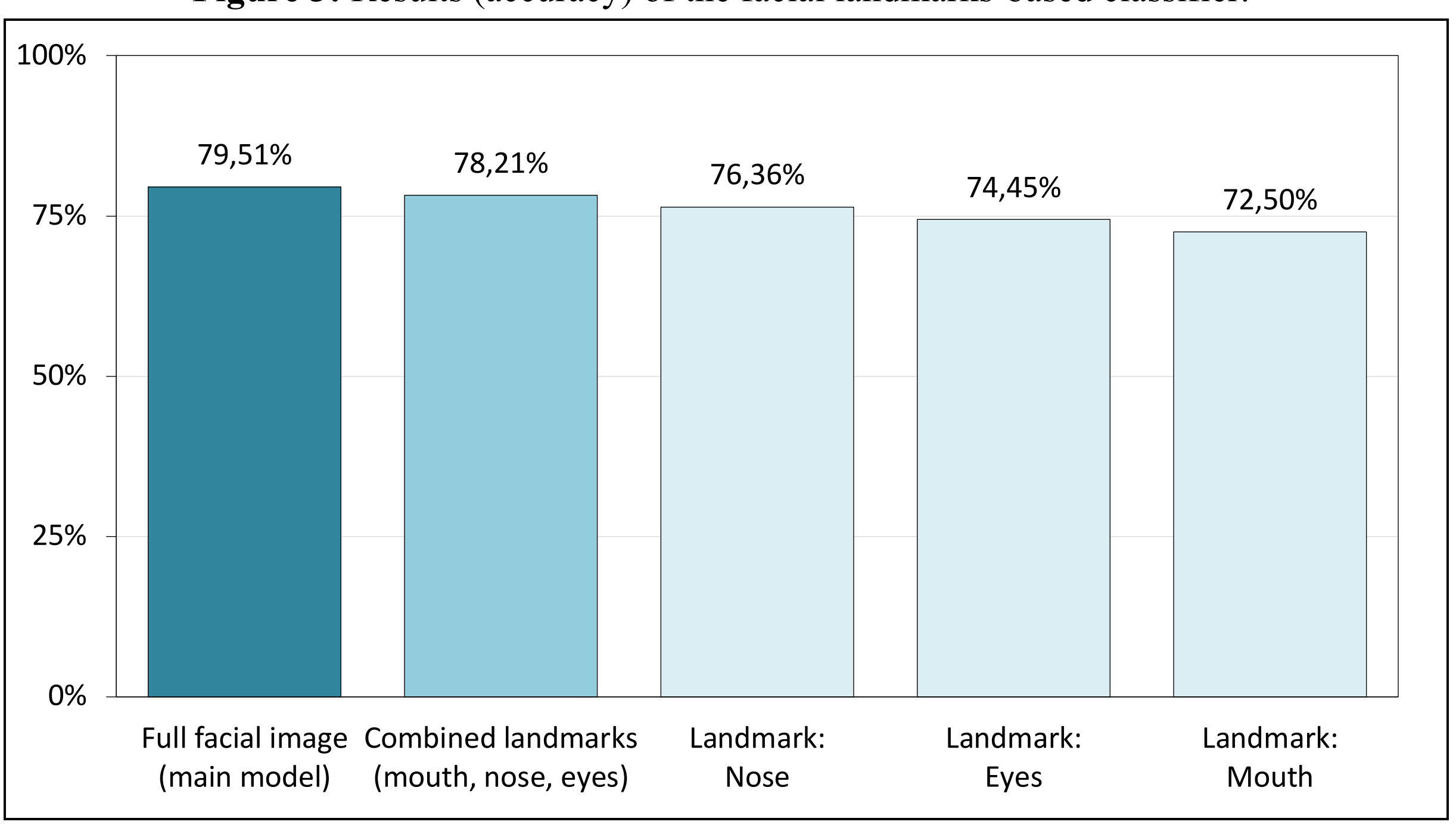

*Notes:* 50% is the accuracy of chance.

**Figure 4:** Accuracy of our AI model vs. human experts and trained humans.

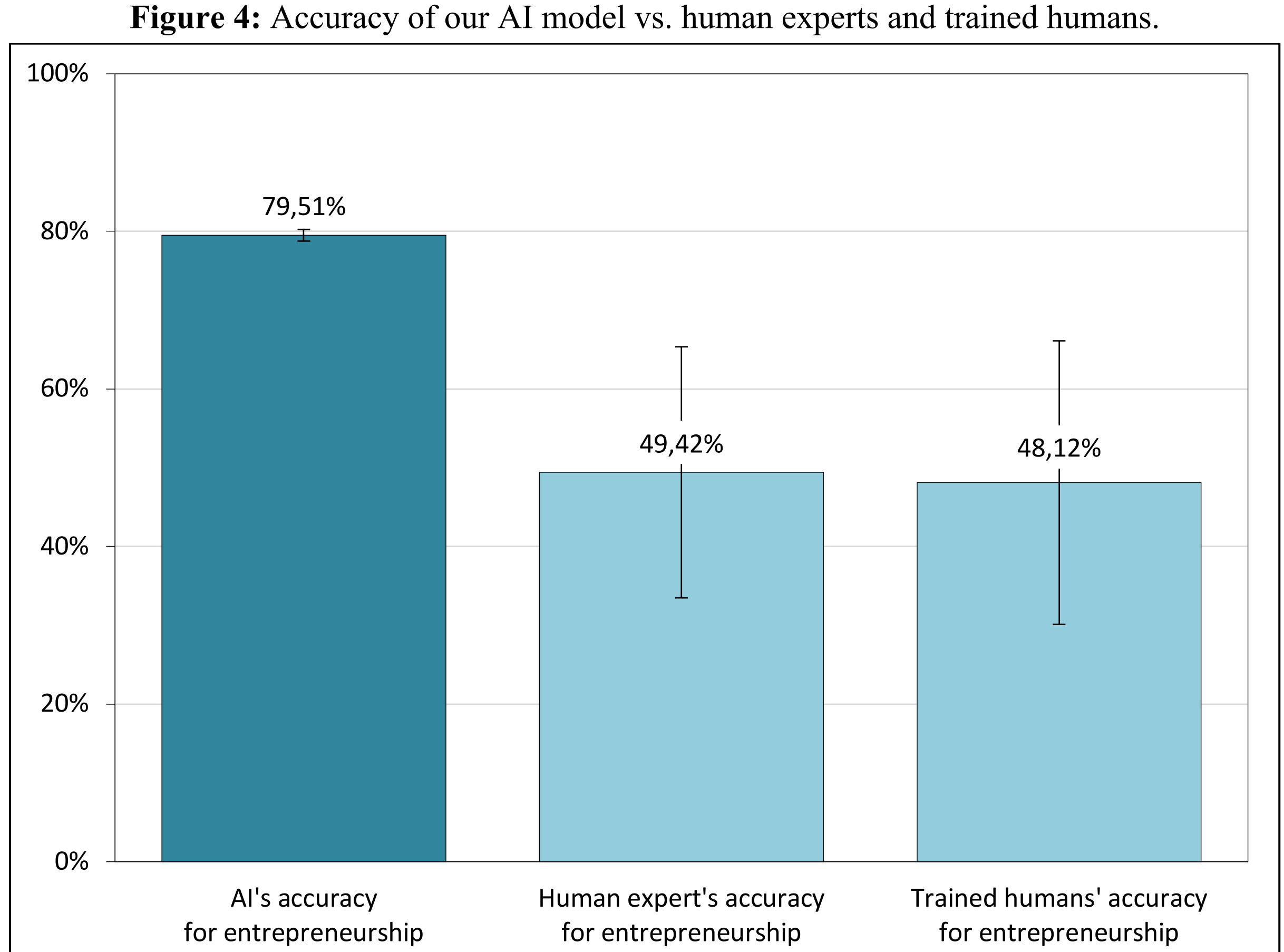

## Appendix A: Technical details, further analyses, and robustness checks

*This appendix provides additional technical details regarding our method and analyses. To keep the appendix more self-contained, we repeat some parts of the main text verbatim.*

### A.1 Full facial image-based classifier

#### A.1.1 Face detector

The location of the face in the input image is detected using the widely used MTCNN face detection technique which is proposed in Zhang et al. (2016)[1]. Despite being lightweight, this technique allows us to generate real-time inferences. MTCNN achieves competitive results in FDDB (Jain and Learned-Miller, 2010) and WIDER FACE (Yang et al., 2016) benchmarks, outperforming numerous baselines (e.g., Chen et al., 2014; Yang et al., 2015; Yang et al., 2016). This face detection method utilizes three stages of CNNs in its pipeline. In the first stage, candidate windows for the face region are produced through a proposal network. These proposals are fed to another CNN (Refine Network) which further rejects the false candidates. In the final stage, the bounding box is properly aligned to the face by detecting five facial landmarks' positions, namely, the eyes, nose, and the corners of the mouth. Utilizing this framework, we detect the face region of each input image. Then, the detected faces are resized to 224×224 pixels prior to the facial feature extraction.

#### A.1.2 Extracting facial features using convolutional neural networks

We employ the model proposed in Cao et al. (2018)[2], which is pre-trained using the VGGFace2 dataset as our facial feature extractor. The VGGFace2 architecture (Cao et al., 2018) utilizes ResNet-50 (He et al., 2016), which is a particular CNN structure, as its backbone for facial feature extraction from the input face images. Then, Cao et al. (2018) pass these features through Squeeze-and-Excitation blocks (Hu et al., 2018), which explicitly model the channel-wise relationships within the RGB channels in the input image. Hu et al. (2018) demonstrate that this method has been able to learn improved facial representations in terms of facial features, allowing them to achieve competitive results in the VGGFace2 dataset for both face identification and face verification tasks. Therefore, we utilize this model that was pre-trained on the VGGFace2 dataset and extract features from the 'activation 40' layer, which produces a 14×14×1024 feature vector when presented with an input face image.

[1] We use the following implementation: https://github.com/ipazc/mtcnn
[2] We use the following implementation: https://github.com/WeidiXie/Keras-VGGFace2-ResNet50

### A.1.3 Entrepreneurship classifier

Convolutional Neural Networks (CNNs) achieve tremendous success in a variety of face-related tasks, including, face recognition (Masi et al., 2018), fake face generation (Gao et al., 2018), and fake face detection (Hsu et al., 2018). For instance, in face verifications, where a candidate image is compared against a target image to verify whether the two face images are coming from the same person or not, CNNs outperform humans (Shepley, 2019). Furthermore, CNNs have been able to generate realistic-looking synthetic faces which are able to fool humans (Hao, 2019). Considering the recent success of CNNs we propose to utilize a classifier based on CNNs.

Our AI model (visually illustrated in Figure 1) is composed of two identical shared feature processing blocks, each with two convolution 2D layers, two activation layers, a batch normalization layer, and a dropout layer. The shared feature processing blocks allow our classifier to seamlessly compare the features from left and right images, without adding a significant parameter overhead.

Each convolution 2D layer has a 3×3 convolution layer with 32 filters. The dropout rate of the respective dropout layers (Srivastava et al., 2014) is indicated within brackets. Subsequently, these features are passed through a dense layer with 512 and Relu activation and a dropout layer with a 0.5 dropout rate. The output is fed into a dense layer with sigmoid activation to generate the binary classification; class 0 if the left image is of an entrepreneur or class 1 if the right image is of an entrepreneur. This classifier is trained using binary cross-entropy loss and Adam (Kingma and Ba, 2014) optimizer with a learning rate of 0.001 and a decay of $1\times10^{-6}$ for 100 epochs.

## A.2 Facial landmarks-based classifier

### A.2.1 Detection of facial landmarks

First the input facial images (after detecting the faces) are passed through a popular facial landmark detection algorithm named face alignment network which is proposed in Bulat and Tzimiropoulos (2017). This architecture is constructed using Hour-Glass network architecture (Newell et al., 2016), where the authors stack two hourglass modules for processing spatial information in multiple scales for dense prediction. This architecture has been able to achieve competitive results for 2D and 3D face alignment and landmark detection in multiple benchmarks, including the AFLW2000-3D dataset (Zhu et al., 2016).

As the facial landmarks, we use the eyes, nose, and mouth regions. After detecting these regions rest of the face is masked out such that it contains information only for a particular landmark region and the rest of the pixels of the image are composed of zeros. Figure A2 provides examples of nose, eyes, and mouth regions.

### A.2.2 Feature extraction

To draw a direct comparison between the two experiments we apply the same feature extraction strategy in Section where we pass the generated images through the VGG-Face2 model and extract facial features from the 'activation 40' layer.

### A.2.3 Classifier

Once the non-facial landmark regions are masked out only a smaller region within the extracted feature vector contains informative details. Therefore, a shallow (with a smaller number of trainable parameters) can be utilized compared to our initial entrepreneurship classifier which receives the full facial image.

This model consists of two modules each with a 3×3 convolutional layer with 64 filters followed by batch normalization, a Relu nonlinearity, and a dropout layer with a 0.25 dropout rate. Then we pass the resultant feature vector through a max pooling layer with 2×2 pool size and the final entrepreneurship classification is generated by a dense layer with sigmoid activation. This classifier is trained using binary cross-entropy loss and Adam (Kingma and Ba, 2014) optimizer with a learning rate of 0.001 and a decay of $1\times10^{6}$ for 100 epochs.

For each landmark (eye, nose, and mouth), 75% of the data is randomly chosen for training and 25% of the data for testing, and a separate model is trained for each land- mark. We repeated this process three times and the average accuracy on the test set is chosen as the final accuracy. We also evaluate the accuracy of combining these three landmarks. In this model, we first pass the individual landmarks through two convolution- batch-normalization - Relu-nonlinearity - dropout modules as defined previously and then through a 2×2 max pooling layer. Then we concatenate the outputs from the individual landmark streams and generate the final classification by a dense layer with sigmoid activation.

## A.3 Further analysis regarding gender

Deep learning models learns task specific features automatically. These features are represented in a model specific, higher dimensional embedding space. However, this embedding space is not human interpretable, and we need to use embedding space visualization techniques make it interpretable. Therefore, to better understand what features of the face region are extracted by the face classifier and to understand whether these identified features have any bias towards a specific gender, we generate an embedding space visualization for a set of randomly chosen samples. Our VGG-Face 2 feature extractor extracts features from each facial image that is in the given pair of images which represent the important features in that particular face. This extracted feature vector is of size 512, as such, it can be seen as representing the 224×224 = 50,176 pixels in the input face image in 512 values. Therefore, each image in our training set of face images is represented with 512 values and we call this space as our embedding space. The embedding space is the learned representation space where each image is represented based on its salient characteristics. To visualize this embedding space, we need to map the 512 dimensions either into 2 dimensions (2D) or into 3 dimensions using a dimensionality reduction scheme such as principal component analysis. To analyze whether the grouping (or clustering) of the samples in the embedding space is based on gender or whether it's governed by the learned entrepreneurship characteristics we generated a 2D embedding space visualization for 250 male and 60 female samples that were randomly chosen from the test set.

In both categories, 50% of the sample are facial images from entrepreneur (ENT) and the remaining 50% are from non-entrepreneurs (Non-ENTs). We apply the PCA algorithm (Wold et al., 1987) on the output of the feature extractor to visualize the embeddings in 2D. The resulting plot is displayed in Figure A3. This analysis shows that the AI model's learned embedding space separation is based on the ENT/Non-ENT labels and does not seem to be biased towards a specific gender. We would like to note that the examples of the same color (ENT: pink, Non-ENT: green) are clustered together irrespective of the shape (circles-males, stars-females). This shows that the proposed AI model has been able to automatically retrieve ENT-specific features from the input faces irrespective of their gender.

### A. References

## A. Figures

**Figure A1:** Training and validation accuracy curves, suggesting that overfitting is not a major concern.

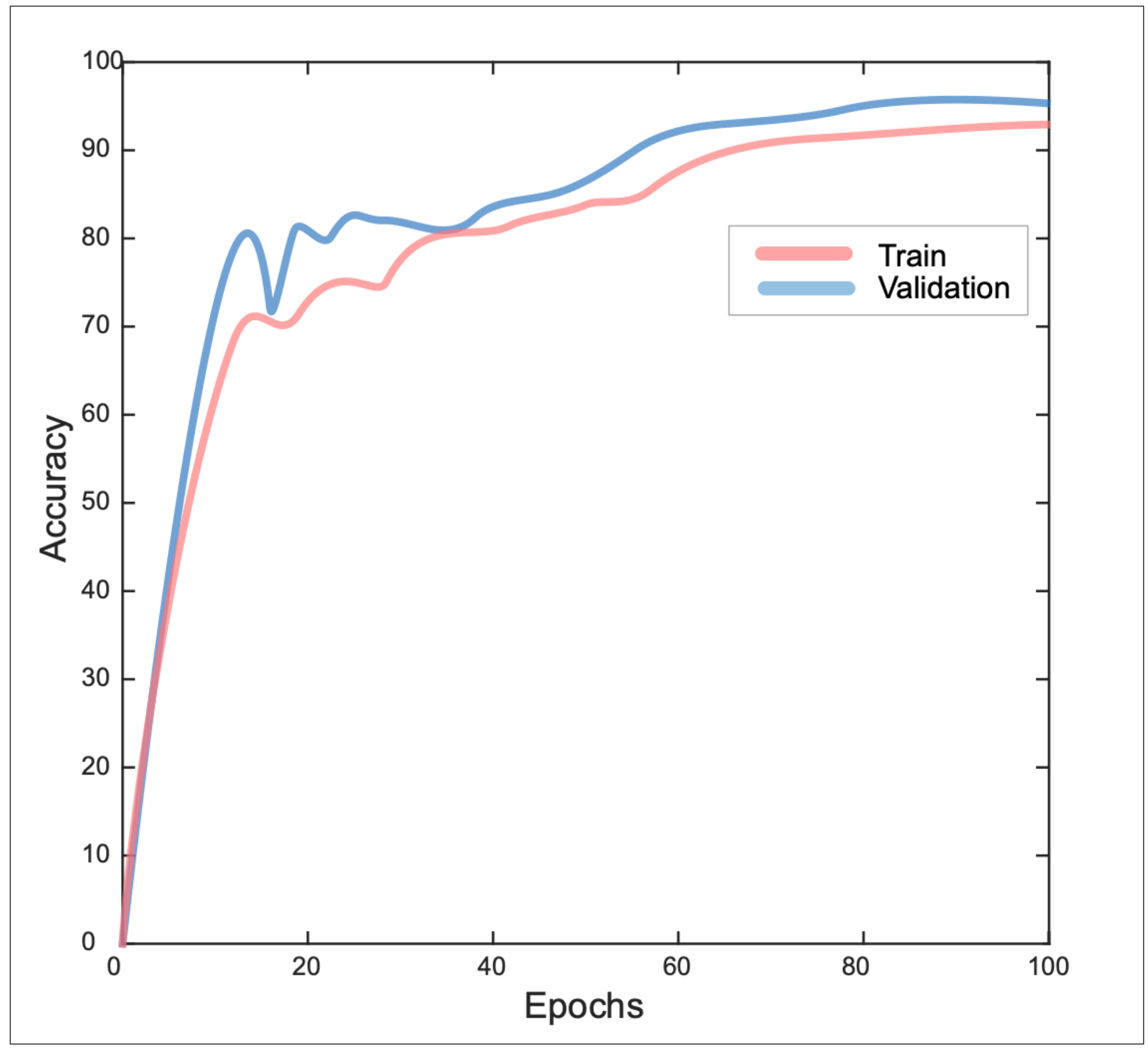

**Figure A2:** Examples for the landmark images.

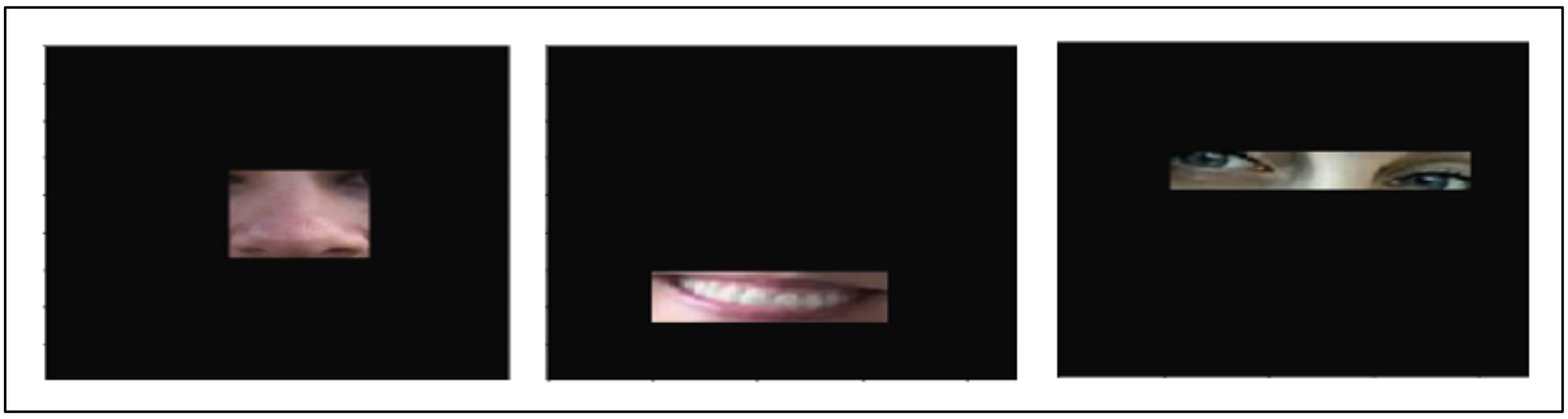

**Figure A3:** Model bias and sensitivity analysis.

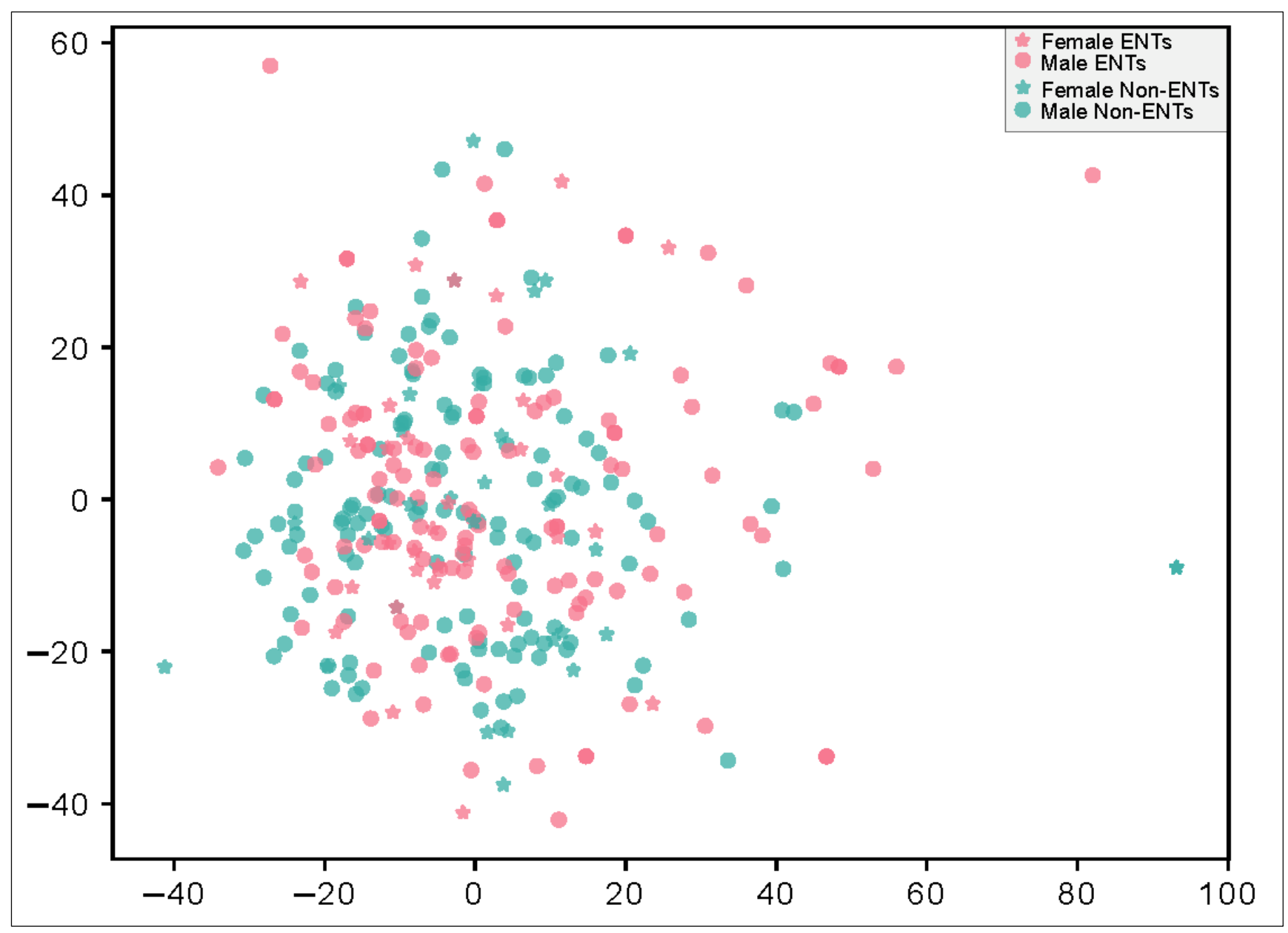

*Notes*: 2D embedding space visualization showing a randomly chosen 250 male and 60 female samples from the test set. In both categories, 50% of the samples are ground truth entrepreneur (ENT) faces and the remaining 50% are from non-entrepreneurs (Non-ENTs). Both male and female ENTs are clustered together irrespective of gender. Similarly, Non-ENTs are also clustered together.